\pdfoutput=1
\documentclass[letterpaper]{article} 
\usepackage{aaai25}  
\usepackage{times}  
\usepackage{helvet}  
\usepackage{courier}  
\usepackage[hyphens]{url}  
\usepackage{graphicx} 
\usepackage{natbib}  
\usepackage{caption} 
\usepackage{algorithm}
\usepackage{algorithmic}

\usepackage{amsmath}
\usepackage{multirow}
\usepackage[table,xcdraw]{xcolor}

\usepackage{newfloat}
\usepackage{listings}
\DeclareCaptionStyle{ruled}{labelfont=normalfont,labelsep=colon,strut=off} 
\floatstyle{ruled}
\newfloat{listing}{tb}{lst}{}
\floatname{listing}{Listing}
\nocopyright

\title{Rare Event Detection in Imbalanced Multi-Class Datasets Using an Optimal MIP-Based Ensemble Weighting Approach}
\author{
    Georgios Tertytchny\textsuperscript{\rm 1,2}\thanks{Corresponding author.}\equalcontrib, Georgios L. Stavrinides\textsuperscript{\rm 1}\equalcontrib, Maria K. Michael\textsuperscript{\rm 1,2}\\
}
\affiliations{
    
    \textsuperscript{\rm 1}KIOS Research and Innovation Center of Excellence\\
    \textsuperscript{\rm 2}Department of Electrical and Computer Engineering, University of Cyprus\\
    \{tertytchny.georgios, stavrinides.georgios, mmichael\}@ucy.ac.cy\\
}

\begin{document}

\maketitle

\begin{abstract}

To address the challenges of imbalanced multi-class datasets typically used for rare event detection in critical cyber-physical systems, we propose an optimal, efficient, and adaptable mixed integer programming (MIP) ensemble weighting scheme. 
Our approach leverages the diverse capabilities of the classifier ensemble on a granular per class basis, while optimizing the weights of classifier-class pairs using elastic net regularization for improved robustness and generalization.
Additionally, it seamlessly and optimally selects a predefined number of classifiers from a given set.
We evaluate and compare our MIP-based method against six well-established weighting schemes, using representative datasets and suitable metrics, under various ensemble sizes.
The experimental results reveal that MIP outperforms all existing approaches, achieving an improvement in balanced accuracy ranging from 0.99\% to 7.31\%, with an overall average of 4.53\% across all datasets and ensemble sizes. 
Furthermore, it attains an overall average increase of 4.63\%, 4.60\%, and 4.61\% in macro-averaged precision, recall, and F1-score, respectively, while maintaining computational efficiency.

\end{abstract}

%
\begin{links}
     \link{Code}{https://github.com/gterty019/MIPENS}
\end{links}

\section{Introduction}
\label{sec:intro}

Rare event detection in cyber-physical systems (CPS) is vital for maintaining the safety and reliability of critical infrastructures, such as water distribution networks and power grids \cite{Taheri2024, Tertytchny2020}.
Critical CPS often generate imbalanced multi-class datasets where some classes have significantly fewer instances than others, due to the infrequent occurrence of abnormal events (e.g., faults or attacks) compared to normal activities \cite{Yin2020}. 
Class imbalance poses several challenges, such as biased model performance towards the majority classes, poor generalization to minority classes, and thus increased difficulty in detecting rare events \cite{Shyalika2023}.
These issues are further amplified by the critical nature of CPS, where misclassifying low-probability but high-impact events may lead to safety risks or system failures. 
Consequently, addressing these challenges is crucial for effective rare event detection.

Ensemble learning approaches, especially weighted voting techniques, are commonly employed to handle imbalanced multi-class datasets \cite{Khan2024}. 
The aim of such methods is to improve overall performance and robustness, by leveraging the strengths of diverse classification algorithms.
Specifically, a weighted voting ensemble model combines the predictions of multiple classifiers, using different weights based on their predictive performance. Higher weights are assigned to better-performing classifiers, either at an overall classifier level or on a per class basis \cite{Dogan2019}.
However, existing weighting schemes often fail to optimally balance the contribution of each classifier, especially in the presence of a highly skewed class distribution.
Moreover, since smaller ensembles are typically less computationally demanding, and given that rare event detection in CPS is often performed at the network edge on resource-constrained devices, adaptable approaches are required to seamlessly and optimally select a predefined number of classifiers from a larger set while assigning their weights \cite{Danso2022}.

Aiming to fill this gap, we propose an optimal, adaptable, and efficient mixed integer programming (MIP) weighting scheme for weighted voting ensembles, suitable for rare event detection in imbalanced multi-class datasets. 
Our approach optimally assigns weights to classifiers per class based on their performance, leveraging their individual capabilities while selecting the predefined number of classifiers from a given set. 
To prevent overfitting and enhance generalization in weight assignment, we incorporate into our method the elastic net regularization technique \cite{Zou2005}, applying it beyond its traditional use in the training and validation phase.
The proposed weighting scheme aims to enhance performance across all classes while ensuring computational efficiency, effectively addressing the limitations of existing approaches.
Our method optimizes weights based on empirical accuracy results, avoiding simplifying assumptions typically required by theoretical analysis, such as independent and identically distributed instances, which may not hold in realistic scenarios involving class imbalance and diverse classifier behavior \cite{Opitz1999}.

Our main contributions are summarized as follows:
\begin{itemize}
    \item We propose an optimal MIP-based ensemble weighting approach for imbalanced multi-class datasets to improve rare event detection in critical CPS. Our weighting scheme optimizes classifier weights on a per class basis, leveraging the diverse strengths of each classifier in a granular manner. 

    \item The proposed method seamlessly and optimally selects a predefined number of classifiers from a given set while calculating their weights. Moreover, it utilizes elastic net regularization to improve generalization and robustness in the assignment of classifier weights, while maintaining computational efficiency.

    \item We demonstrate the improved performance, adaptability, and efficiency achieved by our approach through comparative evaluation against six well-established weighting schemes, using representative datasets and appropriate metrics under different ensemble sizes.
\end{itemize}

To our knowledge, in contrast to our approach, none of the existing ensemble weighting schemes are intrinsically adaptable to varied ensemble sizes and capable of optimally assigning classifier weights for imbalanced multi-class datasets, while attaining generalization and efficiency.


\section{Related Work}
\label{sec:BackgroundRelatedWork}

Weighted voting ensemble methods typically apply weights per classifier or per classifier and class, based on classifier performance in the training and validation phase \cite{Maheshwari2022}. 
Among the fundamental techniques, UW-PC (uniform weights per classifier) and UW-PCC (uniform weights per classifier and class) assign equal weights to all classifiers and classifier-class pairs, respectively, and commonly serve as baselines in comparative studies \cite{Dong2020}. 
On the other hand, WA-PC (weighted average based on normalized accuracy
per classifier) and WA-PCC (weighted average based on normalized accuracy per classifier and class) apply weights based on the normalized accuracy of each classifier or classifier-class pair, respectively \cite{Sagi2018}. Both techniques assign greater weights to classifiers or classifier-class pairs that exhibit better performance. 
DE (differential evolution) is an evolutionary algorithm that determines classifier weights in an iterative refining manner \cite{Storn1997, Zhang2014, Zhang2024}. 
In contrast, BMA (Bayesian model averaging) assigns weights to classifier-class pairs based on their Bayesian inference posterior probabilities, which reflect the likelihood that each classifier is accurate given the observed data \cite{Raftery1997, Wang2022}.

However, these weighting schemes do not optimally assign classifier weights, are not easily adaptable to a varied number of classifiers, and fail to address highly uneven class distributions while achieving generalization.
In contrast, our MIP approach attains generalization in weight calculation through elastic net regularization, which is a regularized regression technique. It linearly combines the widely used L1 and L2 penalties of the lasso and ridge methods, respectively, to leverage the advantages of both strategies \cite{Zou2005}. This dual technique can lead to a weight assignment that is both sparse (less likely to overfit) and robust (less sensitive to training data noise) \cite{Giesen2019}.

Previous studies have explored different perspectives and strategies to address the challenges of imbalanced datasets. 
A comprehensive review of the problem, potential solutions, and applicable evaluation metrics are examined in \cite{He2009}. 
A general evaluation framework is proposed in \cite{Khan2024}, employing data augmentation and ensemble techniques by analyzing binary datasets with varied class imbalance.
In \cite{Lango2022}, the impact of multi-class imbalanced data difficulty factors (such as class overlapping and skewed distribution) on classifier performance is examined.
On the other hand, two approaches integrating sampling, data space improvement, ensemble, and self-paced learning are introduced in \cite{Liu2020} and \cite{Yin2020} to handle class imbalance. However, both methods are tailored to binary classification.
Overall, previous studies lack an optimal and adaptable technique for classifier weight assignment to enhance ensemble performance in imbalanced multi-class datasets, while achieving generalization and computational efficiency across varied ensemble sizes.

\section{Proposed Weighting Scheme}
\label{sec:ProposedWeightingScheme}

\begin{figure}[t]
    \centering
    \includegraphics[width=0.93\columnwidth]{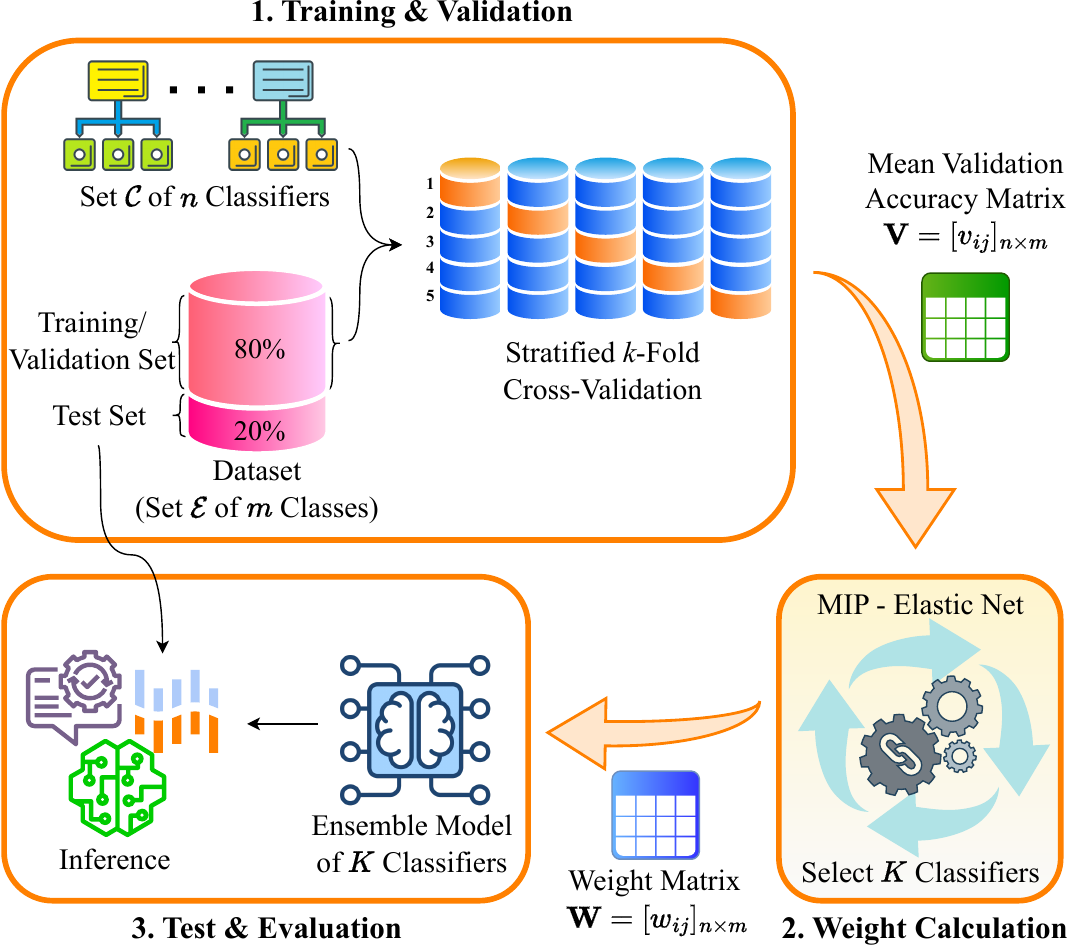}
    \caption{Overview of methodology for incorporating the proposed weighting scheme into the ensemble model.} 
    \label{fig:proposedws}
\end{figure}

To incorporate the proposed weighting approach into the ensemble model, we employ the methodology outlined in Figure \ref{fig:proposedws}. 
Specifically, given a set of $n$ classifiers $\mathcal{C} = \{ C_1, C_2, \dots , C_n\}$, a set of $m$ classes $\mathcal{E} = \{E_1, E_2, \dots , E_{m}\}$, and a predefined ensemble size $K$, our methodology consists of three phases:
\begin{enumerate}
    \item Training \& Validation: The input dataset corresponding to $\mathcal{E}$ is used to train each classifier in $\mathcal{C}$, using stratified $k$-fold cross-validation to ensure a similar class distribution in each fold. This yields the mean validation accuracy matrix $\mathbf{V} = [v_{ij}]_{n \times m}$, where $v_{ij}$ denotes the mean validation accuracy of classifier $C_i$ for class $E_j$ across all folds. 
    It is noted that $\mathcal{E}_{\mathrm{N}} \cup \mathcal{E}_{\mathrm{A}} = \mathcal{E}$ and $\mathcal{E}_{\mathrm{N}} \cap \mathcal{E}_{\mathrm{A}} = \emptyset$, where $\mathcal{E}_{\mathrm{N}} \subset \mathcal{E}$ and $\mathcal{E}_{\mathrm{A}} \subset \mathcal{E}$ represent the normal and abnormal (e.g., faults/attacks) classes, respectively, in $\mathcal{E}$. 

    \item Weight Calculation: We use $\mathbf{V}$ to determine the weights for each classifier-class pair. The proposed approach utilizes MIP and elastic net regularization to select the predefined number of classifiers $K$ from the initial set $\mathcal{C}$, while optimally calculating their weights per class. This results in a weight matrix $\mathbf{W} = [w_{ij}]_{n \times m}$, where $w_{ij}$ denotes the weight of classifier $C_i$ for class $E_j$. If a classifier $C_i$ is not selected, then $w_{ij} = 0 \, \, \forall \, E_j \in \mathcal{E}$.
    

    \item Test \& Evaluation: The calculated weights in $\mathbf{W}$ are assigned to the selected $K$ classifiers to evaluate the performance of the resulting ensemble model.
\end{enumerate}
The main problem addressed---determining the optimal class-based weight assignment for classifiers in a weighted voting ensemble model---corresponds to phase 2.

\subsection{Problem Formulation}
\label{subsec:formulation}
To solve the examined problem in phase 2, we formulate it as a MIP model by defining its decision variables, objective function, and constraints.


\subsubsection{Decision Variables.}
\label{subsubsec:variables}
The following variables are used to select $K$ classifiers and to determine their optimal weights:
\begin{itemize}
	\item $x_i$: binary variable corresponding to classifier $C_i$, such that $x_i=1$ if  $C_i$ is selected, and $x_i=0$ otherwise ($\mathbf{X} = [x_i]_{n \times 1}$).
	\item $w_{ij}$: non-negative continuous variable representing the $(i,j)$-th element of weight matrix $\mathbf{W}$, denoting the weight of classifier-class pair $(C_i,E_j)$.
\end{itemize}

\subsubsection{Objective.}
\label{subsubsec:objective}
We aim to maximize the overall classification accuracy of the ensemble while ensuring generalization, i.e.:
\begin{equation}
\label{eq:objective}
\begin{split}
\max_{\mathbf{X}, \mathbf{W}} & \left[ \frac{1}{m} \sum_{C_i \in \mathcal{C}} \sum_{E_j \in \mathcal{E}} w_{ij}\,v_{ij} 
\right.\\
& \,\,\, - \left. \lambda \left(\alpha \sum_{C_i \in \mathcal{C}} \sum_{E_j \in \mathcal{E}} w_{ij}
+ \frac{1-\alpha}{2} \sum_{C_i \in \mathcal{C}} \sum_{E_j \in \mathcal{E}} w_{ij}^2 \right) \right]
\end{split}
\end{equation}
subject to the constraints defined in the next subsection.

In \eqref{eq:objective}, the overall accuracy is denoted by the first term, where $v_{ij}$ represents the $(i,j)$-th element of mean validation accuracy matrix $\mathbf{V}$, denoting the mean accuracy of classifier $C_i$ for class $E_j$ in the training and validation phase (phase 1).
To prevent the ensemble model from overly relying on specific classifier-class combinations based on the accuracy attained in phase 1, and thus achieve a more balanced and generalizable model, we incorporate into our objective the elastic net regularization technique, represented by the second term in \eqref{eq:objective}.
The L1 and L2 penalties are denoted by $\sum_{C_i \in \mathcal{C}} \sum_{E_j \in \mathcal{E}} w_{ij}$ and $\sum_{C_i \in \mathcal{C}} \sum_{E_j \in \mathcal{E}} w_{ij}^2$, respectively.
The L1 penalty promotes sparsity by driving some weights to zero, while the L2 penalty results in smaller and more evenly distributed weights. Consequently, the combination of both penalties leads to a sparse and robust weight assignment.
In \eqref{eq:objective}, $\lambda$ is the regularization parameter that controls the strength of the overall penalty applied to the objective function ($\lambda \geq 0$). On the other hand, $\alpha$ is the mixing parameter that balances the contribution of the L1 and L2 penalties in the elastic net regularization ($0 \leq \alpha \leq 1$).

\subsubsection{Constraints.}
\label{subsubsec:constraints}
Our objective function \eqref{eq:objective} is solved subject to the following constraints:
\begin{equation}
\label{eq:1}
	x_i \in \{0,1\}, \, \forall \, C_i \in \mathcal{C},	
\end{equation}
\begin{equation}
\label{eq:2}
	w_{ij} \geq 0, \, \forall \, C_i \in \mathcal{C}, \, \forall \, E_j \in \mathcal{E},
\end{equation}
\begin{equation}
	\label{eq:4}
	\sum_{C_i \in \mathcal{C}} x_i = K,
\end{equation} 
\begin{equation}
	\label{eq:5}
	\sum_{C_i \in \mathcal{C}} w_{ij} =1, \, \forall \, E_j \in \mathcal{E},  
\end{equation}
\begin{equation}
\label{eq:6}
	\sum_{E_j \in \mathcal{E}} w_{ij} \leq m\,x_i, \, \forall \, C_i \in \mathcal{C}, 
\end{equation}
\begin{equation}
\label{eq:7}
	\sum_{E_j \in \mathcal{E}} w_{ij} + M(1-x_i) \geq \epsilon, \, \forall \, C_i \in \mathcal{C}, 
\end{equation}
\begin{equation}
\label{eq:8}
	\sum_{C_i \in \mathcal{C}} w_{ij}\,v_{ij} \geq \frac{1}{n}\sum_{C_i \in \mathcal{C}} v_{ij} + \epsilon, \, \forall \, E_j \in \mathcal{E},
\end{equation}
\begin{equation}
\label{eq:9}
	\frac{1}{m} \sum_{C_i \in \mathcal{C}} \sum_{E_j \in \mathcal{E}} w_{ij}\,v_{ij} \geq \frac{1}{n \, m} \sum_{C_i \in \mathcal{C}} \sum_{E_j \in \mathcal{E}} v_{ij}  + \epsilon.
\end{equation} 

Constraints \eqref{eq:1} and \eqref{eq:2} ensure the binary nature and non-negativity of variables $x_i$ and $w_{ij}$, respectively. 
Constraint \eqref{eq:4} enforces the selection of the predefined number of classifiers $K$.
On the other hand, constraint \eqref{eq:5} guarantees that for each class, the sum of weights of all classifiers is equal to one. 
If a classifier is not selected, \eqref{eq:6} enforces all of its weights across all classes to be equal to zero (otherwise, the sum of its weights is bounded by the total number of classes $m$). 
In contrast, constraint \eqref{eq:7} ensures that if a classifier is selected, at least one of its weights across all classes is greater than zero. In \eqref{eq:7}, $\epsilon$ is a positive constant, sufficiently smaller than the variables and parameters in the MIP model. It is used to convert \eqref{eq:7} to a non-strict inequality, as strict inequalities are not supported in MIP. 

On the other hand, $M$ is a positive constant, sufficiently larger than the variables and parameters in the MIP model. It is used to express the conditional aspect of \eqref{eq:7} in linear form. Specifically, the constraint becomes relevant $\big( \sum_{E_j \in \mathcal{E}} w_{ij} \geq \epsilon \big)$ only if $x_i = 1$. Otherwise, if $x_i = 0$ the constraint becomes irrelevant $\big( \sum_{E_j \in \mathcal{E}} w_{ij} + M \geq \epsilon \big)$, as it is always true.
Constraint \eqref{eq:8} ensures that the selected weights per classifier and class are such that the resulting weighted average accuracy for each class is greater than the average accuracy for the particular class when uniform weights are utilized across all classifiers.
Similarly, \eqref{eq:9} guarantees that the overall weighted average accuracy of the ensemble is greater than the overall average accuracy in the case where all weights of classifier-class pairs are uniform.  
In \eqref{eq:8} and \eqref{eq:9}, $\epsilon$ is utilized in the same manner as in \eqref{eq:7}.

The time complexity of MIP models depends on both the problem size and the solver used. MIP solvers typically employ proprietary algorithms with undisclosed implementation details, and thus their time complexity cannot be easily derived. However, the computational cost of the proposed MIP approach can be considered proportional to the problem size $n \, m$, where $n$ (number of classifiers) is small in practice \cite{Dogan2019}. Consequently, the computational cost grows approximately linearly with respect to $m$ (number of classes). In CPS rare event detection, $m$ is typically small \cite{Vrachimis2018}, as the aim is to distinguish between normal and rare abnormal states rather than to conduct a fine-grained classification. It is noted that our MIP model is independent of the dataset size. These factors ensure that the runtime of our MIP method remains reasonable, as demonstrated in our experimental results.

\section{Experimental Framework}
\label{sec:Experiments}

We evaluated and compared the performance and efficiency of the proposed weighting scheme against six well-established approaches. The comparative evaluation was conducted using appropriate metrics and four relevant imbalanced multi-class datasets for rare event detection in critical CPS, under varied ensemble sizes.

\subsection{Experimental Setup}
\label{subsec:ExperimentalSetup}
The three phases of our methodology were implemented using a server running CentOS 7.9, equipped with an Intel Xeon Gold 6240 processor @ 2.6\,GHz and 400\,GB of RAM. The software environment included Python 3.12 with scipy 1.14.0, scikit-learn 1.5.1, NumPy 2.0, pandas 2.2.2, and Gurobi Optimizer 11.0.3 with its Python interface gurobipy. We utilized the Gurobi solver in phase 2 to solve our MIP-formulated model, as it is widely used to optimally solve similar optimization problems \cite{Gurobi}.
We split each examined dataset into 80\% for training and validation, used in phase 1, and 20\% for the test set, used in phase 3. 
In phase 1, we employed stratified $k$-fold cross-validation with $k=5$ \cite{McTavish2022}.
To fine-tune parameters $\lambda$ and $\alpha$ in \eqref{eq:objective}, we incremented/decremented each parameter, terminating adjustments when performance began to deteriorate. This fine-tuning resulted in the following values for $(\lambda, \alpha)$ for each of the four examined datasets: $(0.95, 0.85)$, $(0.96, 0.80)$, $(1.00, 0.82)$, and $(0.95, 0.86)$, respectively.

\subsection{Existing Weighting Schemes}
\label{subsec:existingSchemes}
We compared our MIP approach against the six widely used weighting methods discussed in related work: UW-PC, UW-PCC, WA-PC, WA-PCC, DE, and BMA.
In our experiments, UW-PC and UW-PCC served as baselines. 
Without loss of generality, we used the mean validation accuracy matrix $\mathbf{V}$ as the likelihood matrix in BMA \cite{Zhou2021}.

\subsection{Classifiers}
We considered a set of $n = 8$ commonly used classifiers $\mathcal{C}$ $=\{$MLR, J48, JRIP, REPTree, MLP, SVM, GNB, IBk$\}$. These classifiers were selected for their complementary strengths in multi-class classification, as they encompass a diverse range of approaches: probabilistic (GNB), rule-based (JRIP), tree-based (J48, REPTree), function-based (MLR, MLP, SVM), and instance-based (IBk) \cite{Hastie2009}. All classifiers are available in the scikit-learn 1.5.1 Python library.
In our experiments, we used ensemble sizes ranging from $K = 2$ to $K = 8$ to investigate a comprehensive set of scenarios.

\begin{table}[t]
    \centering
    \fontsize{9}{10}\selectfont
    \begin{tabular}{llrr}
        \hline
        $E_j$ & \textbf{Class Description} & \multicolumn{1}{c}{\textbf{\# Instances}} & \multicolumn{1}{r}{\textbf{\%}} \\
        \hline

        \multicolumn{4}{c}{{\textbf{Dataset D1:} LeakDB ($\rho = 6059.97$)}} \\ \hline
        $N_1$ & Normal - No Leak/No Pressure   & 17\,520 & 3.14 \\ 
        $N_2$ & Normal - No Leak/High Pressure & 448\,438 & 80.38 \\ 
        $N_3$ & Normal - No Leak/Low Pressure  & 88\,154 & 15.80 \\ 
        $F_1$ & Abrupt - Large Leak/High Pressure & 1\,721 & 0.31 \\ 
        $F_2$ & Incipient - Large Leak/High Pressure & 1\,181 & 0.21 \\ 
        $F_3$ & Incipient - Large Leak/Low Pressure & 812 & 0.15 \\ 
        $F_4$ & Incipient - Small Leak/Low Pressure & 74 & 0.01 \\ \hline
        \multicolumn{2}{l}{Total} & 557\,900 & 100.00 \\ \hline
        
        \multicolumn{4}{c}{{\textbf{Dataset D2:} NSL-KDD ($\rho = 1222.64$)}} \\ \hline
        $N_1$ & Normal & 13\,449 & 53.39 \\
        $A_1$ & Probe & 2\,289 & 9.09 \\
        $A_2$ & Denial of Service (DoS) & 9\,234 & 36.65 \\
        $A_3$ & User to Root (U2R) & 11 & 0.04 \\
        $A_4$ & Remote to Local (R2L) & 209 & 0.83 \\ \hline
        \multicolumn{2}{l}{Total} & 25\,192 & 100.00 \\ \hline
    
        \multicolumn{4}{c}{{\textbf{Dataset D3:} SG-MITM ($\rho = 5.91$)}} \\ \hline
        $N_1$ & Normal & 455 & 65.28 \\ 
        $A_1$ & MITM on Router-Controller & 84 & 12.05 \\ 
        $A_2$ & MITM on Controller & 77 & 11.05 \\ 
        $A_3$ & MITM on Router & 81 & 11.62 \\ \hline
        \multicolumn{2}{l}{Total} & 697 & 100.00 \\ \hline
        
        \multicolumn{4}{c}{{\textbf{Dataset D4:} CIC-IDS2017 ($\rho = 2.13$)}} \\ \hline
        $N_1$ & Normal & 273\,097 & 34.52 \\ 
        $A_1$ & Distributed Denial of Service (DDoS) & 128\,027 & 16.18 \\ 
        $A_2$ & Denial of Service Hulk (DoS Hulk) & 231\,073 & 29.20 \\ 
        $A_3$ & PortScan & 158\,930 & 20.10 \\ \hline
        \multicolumn{2}{l}{Total} & 791\,127 & 100.00 \\ \hline
    \end{tabular}
    \caption{Class distribution in datasets D1--D4 ($N_j \in \mathcal{E}_{\mathrm{N}}$ and $A_j, F_j \in \mathcal{E}_{\mathrm{A}}$).}
    \label{tab:D1D2D3D4}
\end{table}

\subsection{Datasets}
\label{subsec:Datasets}

\begin{figure*}[t]
\begin{minipage}{\columnwidth}
    \begin{minipage}[t]{\columnwidth}
        \centering
        \includegraphics[width=0.91\columnwidth]{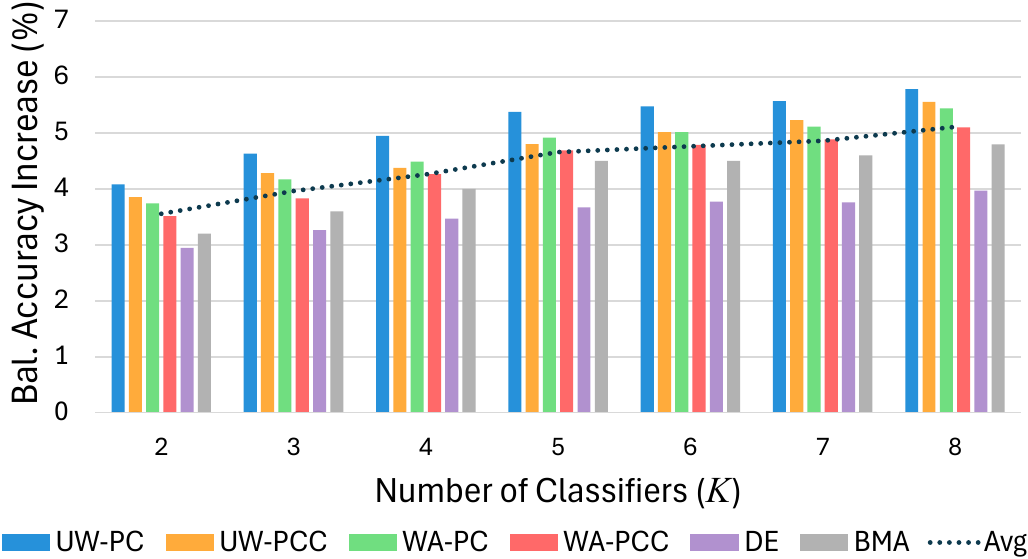}
        \captionof{figure}{Improvement achieved in balanced accuracy by proposed MIP approach over other weighting schemes (D1).}
        \label{fig:D1Improvement}
    \end{minipage}
    \vfill
    \begin{minipage}[t]{\columnwidth}
        \centering
        \begin{table}[H]
            \centering
            \fontsize{9pt}{10pt}\selectfont
                \begin{tabular}{l|l|c|rrrrrrr} 
                \hline
                \multicolumn{3}{r|}{$K$} & \multicolumn{1}{c}{2} & \multicolumn{1}{c}{3} & \multicolumn{1}{c}{4} & \multicolumn{1}{c}{5} & \multicolumn{1}{c}{6} & \multicolumn{1}{c}{7} & \multicolumn{1}{c}{8}\\ \hline
               \multirow{21}{*}{\rotatebox[origin=c]{90}{\textbf{Increase (\%)}}} & \multirow{6}{*}{\rotatebox[origin=c]{90}{\textbf{Precision}}} & UW-PC & $4.06$ & $4.59$ & $4.93$ & $5.34$ & $5.47$ & $5.59$ & $5.88$ \\
               & & UW-PCC & $3.83$ & $4.23$ & $4.45$ & $4.84$ & $4.97$ & $5.21$ & $5.62$ \\
               & & WA-PC & $3.71$ & $4.11$ & $4.57$ & $4.96$ & $4.97$ & $5.09$ & $5.49$ \\
               & & WA-PCC & $3.48$ & $3.76$ & $4.33$ & $4.72$ & $4.85$ & $4.96$ & $5.10$ \\
               & & DE & $2.90$ & $3.17$ & $3.51$ & $3.74$ & $3.74$ & $3.86$ & $4.09$ \\
               & & BMA & $3.48$ & $3.88$ & $4.45$ & $4.96$ & $4.85$ & $5.09$ & $5.23$ \\ \cline{3-10}
               & & \textbf{Avg} & $\mathbf{3.58}$ & $\mathbf{3.96}$ & $\mathbf{4.37}$ & $\mathbf{4.76}$ & $\mathbf{4.81}$ & $\mathbf{4.97}$ & $\mathbf{5.24}$ \\ \cline{2-10}
               & \multirow{6}{*}{\rotatebox[origin=c]{90}{\textbf{Recall}}} & UW-PC & $4.02$ & $4.48$ & $5.04$ & $5.47$ & $5.57$ & $5.67$ & $5.78$ \\
               & & UW-PCC & $3.80$ & $4.25$ & $4.45$ & $4.89$ & $5.11$ & $5.32$ & $5.54$ \\
               & & WA-PC & $3.68$ & $4.13$ & $4.57$ & $5.01$ & $5.11$ & $5.20$ & $5.43$ \\
               & & WA-PCC & $3.46$ & $3.79$ & $4.34$ & $4.77$ & $4.87$ & $4.97$ & $5.08$ \\
               & & DE & $2.90$ & $3.21$ & $3.53$ & $3.74$ & $3.84$ & $3.83$ & $4.04$ \\
               & & BMA & $3.46$ & $3.90$ & $4.45$ & $5.01$ & $4.99$ & $5.09$ & $5.19$ \\  \cline{3-10}
               & & \textbf{Avg} & $\mathbf{3.55}$ & $\mathbf{3.96}$ & $\mathbf{4.40}$ & $\mathbf{4.82}$ & $\mathbf{4.92}$ & $\mathbf{5.01}$ & $\mathbf{5.18}$ \\ \cline{2-10}
               & \multirow{6}{*}{\rotatebox[origin=c]{90}{\textbf{F1-Score}}} & UW-PC & $3.99$ & $4.54$ & $4.99$ & $5.41$ & $5.53$ & $5.64$ & $5.72$ \\
               & & UW-PCC & $3.76$ & $4.19$ & $4.40$ & $4.81$ & $5.05$ & $5.28$ & $5.47$ \\
               & & WA-PC & $3.64$ & $4.07$ & $4.51$ & $4.93$ & $5.05$ & $5.15$ & $5.47$ \\
               & & WA-PCC & $3.41$ & $3.72$ & $4.28$ & $4.69$ & $4.81$ & $4.91$ & $5.10$ \\
               & & DE & $2.96$ & $3.14$ & $3.46$ & $3.74$ & $3.85$ & $3.85$ & $4.02$ \\
               & & BMA & $3.41$ & $3.83$ & $4.40$ & $4.93$ & $4.93$ & $5.03$ & $5.23$ \\  \cline{3-10}
               & & \textbf{Avg} & $\mathbf{3.53}$ & $\mathbf{3.92}$ & $\mathbf{4.34}$ & $\mathbf{4.75}$ & $\mathbf{4.87}$ & $\mathbf{4.98}$ & $\mathbf{5.17}$ \\ \hline
            \end{tabular}
            \caption{Improvement attained in macro-averaged precision, recall, and F1-score by proposed MIP approach over other weighting schemes (D1).}
            \label{tab:D1Improvement}
        \end{table}
    \end{minipage}
\end{minipage}
\hfill
\begin{minipage}{\columnwidth}
    \begin{minipage}[t]{\columnwidth}
        \centering
        \includegraphics[width=0.91\columnwidth]{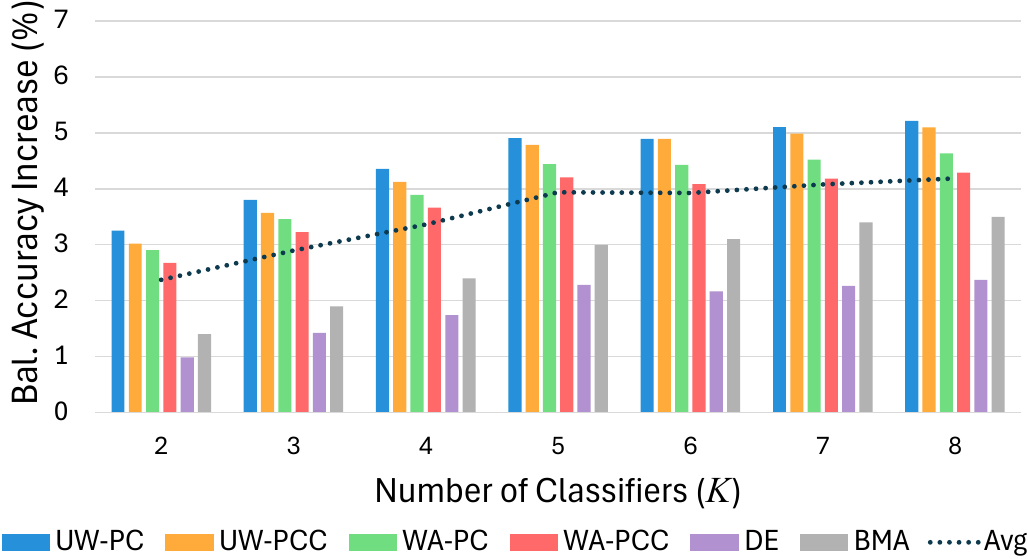}
        \captionof{figure}{Improvement achieved in balanced accuracy by proposed MIP approach over other weighting schemes (D2).}
        \label{fig:D2Improvement}
    \end{minipage}
    \vfill
    \begin{minipage}[t]{\columnwidth}
        \centering
        \begin{table}[H]
            \centering
            \fontsize{9pt}{10pt}\selectfont
                \begin{tabular}{l|l|c|rrrrrrr} 
                \hline
                \multicolumn{3}{r|}{$K$} & \multicolumn{1}{c}{2} & \multicolumn{1}{c}{3} & \multicolumn{1}{c}{4} & \multicolumn{1}{c}{5} & \multicolumn{1}{c}{6} & \multicolumn{1}{c}{7} & \multicolumn{1}{c}{8}\\ \hline
                \multirow{21}{*}{\rotatebox[origin=c]{90}{\textbf{Increase (\%)}}} & \multirow{6}{*}{\rotatebox[origin=c]{90}{\textbf{Precision}}} & UW-PC & $3.25$ & $3.84$ & $4.37$ & $4.87$ & $4.97$ & $5.06$ & $5.04$ \\
                 & & UW-PCC & $3.03$ & $3.62$ & $4.04$ & $4.76$ & $4.97$ & $4.95$ & $4.93$ \\
                 & & WA-PC & $2.92$ & $3.51$ & $3.82$ & $4.43$ & $4.42$ & $4.51$ & $4.49$ \\
                 & & WA-PCC & $2.69$ & $3.29$ & $3.60$ & $4.21$ & $4.09$ & $4.18$ & $4.17$ \\
                 & & DE & $1.06$ & $1.46$ & $1.77$ & $2.17$ & $2.16$ & $2.15$ & $2.15$ \\
                 & & BMA & $1.60$ & $2.10$ & $2.62$ & $3.34$ & $3.44$ & $3.64$ & $3.63$ \\ \cline{3-10}
                 & & \textbf{Avg} & $\mathbf{2.43}$ & $\mathbf{2.97}$ & $\mathbf{3.37}$ & $\mathbf{3.96}$ & $\mathbf{4.01}$ & $\mathbf{4.08}$ & $\mathbf{4.07}$ \\ \cline{2-10}
                 & \multirow{6}{*}{\rotatebox[origin=c]{90}{\textbf{Recall}}} & UW-PC & $3.23$ & $3.79$ & $4.36$ & $4.81$ & $4.80$ & $5.02$ & $5.13$ \\
                 & & UW-PCC & $2.99$ & $3.56$ & $4.12$ & $4.69$ & $4.80$ & $4.90$ & $5.01$ \\
                 & & WA-PC & $2.87$ & $3.44$ & $3.89$ & $4.45$ & $4.32$ & $4.54$ & $4.65$ \\
                 & & WA-PCC & $2.63$ & $3.20$ & $3.65$ & $4.21$ & $4.09$ & $4.19$ & $4.30$ \\
                 & & DE & $1.01$ & $1.46$ & $1.79$ & $2.23$ & $2.12$ & $2.22$ & $2.33$ \\
                 & & BMA & $1.59$ & $2.15$ & $2.71$ & $3.27$ & $3.38$ & $3.72$ & $3.83$ \\  \cline{3-10}
                 & & \textbf{Avg} & $\mathbf{2.39}$ & $\mathbf{2.93}$ & $\mathbf{3.42}$ & $\mathbf{3.94}$ & $\mathbf{3.92}$ & $\mathbf{4.10}$ & $\mathbf{4.21}$ \\ \cline{2-10}
                 & \multirow{6}{*}{\rotatebox[origin=c]{90}{\textbf{F1-Score}}} & UW-PC & $3.24$ & $3.88$ & $4.43$ & $4.85$ & $4.84$ & $5.16$ & $5.15$ \\
                 & & UW-PCC & $3.01$ & $3.65$ & $4.20$ & $4.74$ & $4.84$ & $5.04$ & $5.03$ \\
                 & & WA-PC & $2.90$ & $3.54$ & $3.97$ & $4.39$ & $4.38$ & $4.59$ & $4.58$ \\
                 & & WA-PCC & $2.78$ & $3.31$ & $3.74$ & $4.16$ & $4.04$ & $4.24$ & $4.23$ \\
                 & & DE & $1.09$ & $1.52$ & $1.84$ & $2.15$ & $2.03$ & $2.24$ & $2.24$ \\
                 & & BMA & $1.65$ & $2.18$ & $2.72$ & $3.26$ & $3.36$ & $3.79$ & $3.78$ \\  \cline{3-10}
                 & & \textbf{Avg} & $\mathbf{2.45}$ & $\mathbf{3.01}$ & $\mathbf{3.48}$ & $\mathbf{3.93}$ & $\mathbf{3.92}$ & $\mathbf{4.18}$ & $\mathbf{4.17}$ \\ \hline
            \end{tabular}
            \caption{Improvement attained in macro-averaged precision, recall, and F1-score by proposed MIP approach over other weighting schemes (D2).}
            \label{tab:D2Improvement}
        \end{table}
    \end{minipage}
\end{minipage}
\end{figure*}

We experimented with four representative datasets for rare event detection in critical CPS, featuring multiple classes with varied imbalance: [D1] LeakDB \cite{Vrachimis2018}, [D2] NSL-KDD \cite{Tavallaee2009}, [D3] SG-MITM \cite{Elrawy2023}, and [D4] CIC-IDS2017 \cite{Sharafaldin2018}.
Benchmarks D2 and D4 pertain to intrusion detection in simulated network environments, resembling real-world network traffic captures. They are widely used to investigate anomaly detection in CPS \cite{Zhou2020, Danso2022}. 
On the other hand, D1 involves leakages (faults) in a simulated real-world water distribution network under varying conditions, while D3 focuses on man-in-the-middle (MITM) attacks targeting client-server protocols in the private area network (PAN) of a smart grid CPS, represented by an experimental testbed.
Table \ref{tab:D1D2D3D4} shows the description, number of instances, and the percentage distribution of each normal/abnormal class in datasets D1--D4.
Furthermore, it showcases the imbalance ratio $\rho$ for each dataset, which is defined as the ratio of the number of instances $\mu$ in the majority class to the number of instances $\nu$ in the minority class \cite{Wang2021}. 
Considering that a dataset is highly imbalanced when $\mu \gg \nu$, D1 and D2 are highly imbalanced, whereas D3 and D4 are moderately imbalanced \cite{Liu2020}.

\subsection{Evaluation Metrics}
To provide a comprehensive assessment of our approach against the examined weighting schemes, we employed in test and evaluation phase (phase 3) performance metrics well-suited for imbalanced multi-class datasets.  
Specifically, balanced accuracy served as our primary metric due to its ability to account for class imbalance by averaging the recall obtained for each class. Additionally, we utilized macro-averaged precision, recall, and F1-score as secondary metrics to ensure an unbiased evaluation across all classes, thus providing a holistic comparative analysis \cite{Grandini2020, Lango2022}.

\section{Results \& Discussion}
\label{sec:Results}

\begin{figure*}[t]
\begin{minipage}{\columnwidth}
    \begin{minipage}[b]{\columnwidth}
        \centering
        \includegraphics[width=0.91\columnwidth]{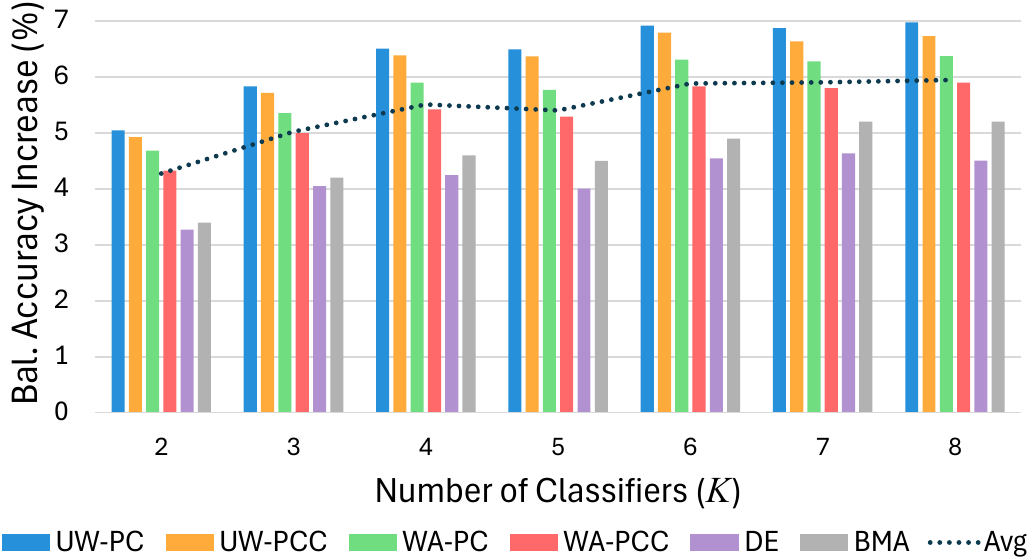}
        \captionof{figure}{Improvement achieved in balanced accuracy by proposed MIP approach over other weighting schemes (D3).}
        \label{fig:D3Improvement}
    \end{minipage}
    \vfill
    \begin{minipage}[b]{\columnwidth}
        \centering
        \begin{table}[H]
            \centering
            \fontsize{9pt}{10pt}\selectfont
                \begin{tabular}{l|l|c|rrrrrrr} 
                \hline
                \multicolumn{3}{r|}{$K$} & \multicolumn{1}{c}{2} & \multicolumn{1}{c}{3} & \multicolumn{1}{c}{4} & \multicolumn{1}{c}{5} & \multicolumn{1}{c}{6} & \multicolumn{1}{c}{7} & \multicolumn{1}{c}{8}\\ \hline
               \multirow{21}{*}{\rotatebox[origin=c]{90}{\textbf{Increase (\%)}}} & \multirow{6}{*}{\rotatebox[origin=c]{90}{\textbf{Precision}}} & UW-PC & $5.02$ & $5.83$ & $6.48$ & $6.46$ & $6.93$ & $6.88$ & $6.99$ \\
                & & UW-PCC & $4.89$ & $5.70$ & $6.35$ & $6.33$ & $6.80$ & $6.63$ & $6.72$ \\
                & & WA-PC & $4.64$ & $5.44$ & $5.83$ & $5.82$ & $6.28$ & $6.37$ & $6.46$ \\
                & & WA-PCC & $4.27$ & $5.06$ & $5.45$ & $5.31$ & $5.77$ & $5.86$ & $5.80$ \\
                & & DE & $3.29$ & $4.18$ & $4.31$ & $3.93$ & $4.51$ & $4.61$ & $4.51$ \\
                & & BMA & $3.90$ & $4.81$ & $5.19$ & $5.05$ & $5.52$ & $5.73$ & $5.80$ \\ \cline{3-10}
                & & \textbf{Avg} & $\mathbf{4.34}$ & $\mathbf{5.17}$ & $\mathbf{5.60}$ & $\mathbf{5.48}$ & $\mathbf{5.97}$ & $\mathbf{6.01}$ & $\mathbf{6.05}$ \\ \cline{2-10}
                & \multirow{6}{*}{\rotatebox[origin=c]{90}{\textbf{Recall}}} & UW-PC & $5.02$ & $5.84$ & $6.42$ & $6.41$ & $6.97$ & $6.82$ & $6.92$ \\
                & & UW-PCC & $4.89$ & $5.71$ & $6.29$ & $6.28$ & $6.85$ & $6.57$ & $6.67$ \\
                & & WA-PC & $4.64$ & $5.34$ & $5.92$ & $5.78$ & $6.35$ & $6.19$ & $6.42$ \\
                & & WA-PCC & $4.27$ & $5.09$ & $5.42$ & $5.28$ & $5.85$ & $5.82$ & $5.80$ \\
                & & DE & $3.29$ & $4.10$ & $4.31$ & $3.94$ & $4.50$ & $4.60$ & $4.47$ \\
                & & BMA & $3.90$ & $4.84$ & $5.17$ & $5.04$ & $5.60$ & $5.70$ & $5.68$ \\  \cline{3-10}
                & & \textbf{Avg} & $\mathbf{4.34}$ & $\mathbf{5.15}$ & $\mathbf{5.59}$ & $\mathbf{5.46}$ & $\mathbf{6.02}$ & $\mathbf{5.95}$ & $\mathbf{5.99}$ \\ \cline{2-10}
                & \multirow{6}{*}{\rotatebox[origin=c]{90}{\textbf{F1-Score}}} & UW-PC & $5.02$ & $5.90$ & $6.51$ & $6.50$ & $6.83$ & $6.91$ & $6.90$ \\
                & & UW-PCC & $4.89$ & $5.77$ & $6.39$ & $6.37$ & $6.70$ & $6.66$ & $6.64$ \\
                & & WA-PC & $4.64$ & $5.39$ & $5.88$ & $5.73$ & $6.32$ & $6.28$ & $6.39$ \\
                & & WA-PCC & $4.27$ & $5.01$ & $5.50$ & $5.36$ & $5.81$ & $5.90$ & $5.75$ \\
                & & DE & $3.29$ & $4.14$ & $4.25$ & $4.00$ & $4.45$ & $4.55$ & $4.37$ \\
                & & BMA & $3.90$ & $4.89$ & $5.24$ & $5.11$ & $5.44$ & $5.78$ & $5.62$ \\  \cline{3-10}
                & & \textbf{Avg} & $\mathbf{4.34}$ & $\mathbf{5.18}$ & $\mathbf{5.63}$ & $\mathbf{5.51}$ & $\mathbf{5.93}$ & $\mathbf{6.01}$ & $\mathbf{5.95}$ \\ \hline
            \end{tabular}
            \caption{Improvement attained in macro-averaged precision, recall, and F1-score by proposed MIP approach over other weighting schemes (D3).}
            \label{tab:D3Improvement}
        \end{table}
    \end{minipage}
\end{minipage}
\hfill
\begin{minipage}{\columnwidth}
    \begin{minipage}[b]{\columnwidth}
        \centering
        \includegraphics[width=0.91\columnwidth]{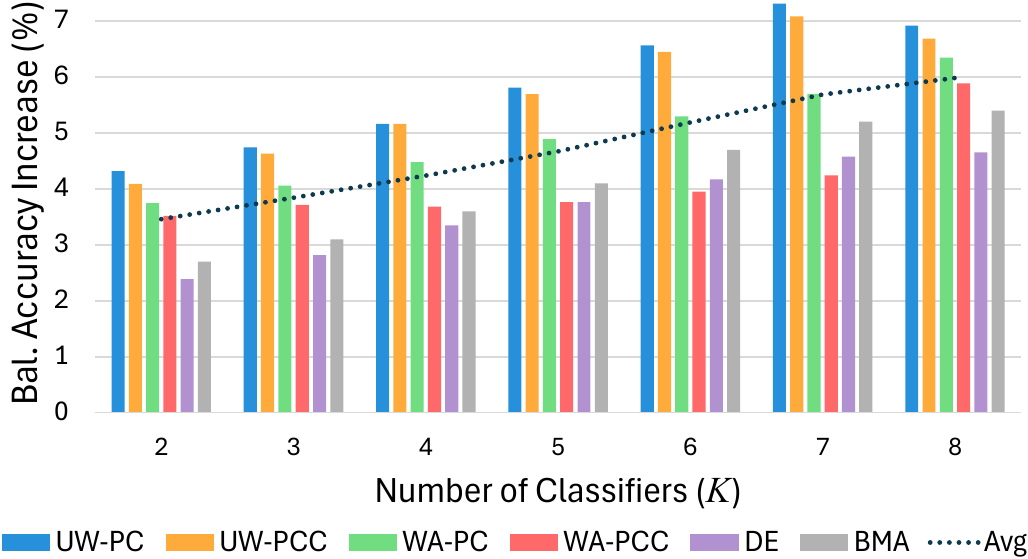}
        \captionof{figure}{Improvement achieved in balanced accuracy by proposed MIP approach over other weighting schemes (D4).}
        \label{fig:D4Improvement}
    \end{minipage}
    \vfill
    \begin{minipage}[b]{\columnwidth}
        \centering
        \begin{table}[H]
            \centering
            \fontsize{9pt}{10pt}\selectfont
                \begin{tabular}{l|l|c|rrrrrrr} 
                \hline
                \multicolumn{3}{r|}{$K$} & \multicolumn{1}{c}{2} & \multicolumn{1}{c}{3} & \multicolumn{1}{c}{4} & \multicolumn{1}{c}{5} & \multicolumn{1}{c}{6} & \multicolumn{1}{c}{7} & \multicolumn{1}{c}{8}\\ \hline
                \multirow{21}{*}{\rotatebox[origin=c]{90}{\textbf{Increase (\%)}}} & \multirow{6}{*}{\rotatebox[origin=c]{90}{\textbf{Precision}}} & UW-PC & $1.68$ & $4.74$ & $5.13$ & $5.84$ & $6.54$ & $7.32$ & $6.89$ \\
                & & UW-PCC & $5.81$ & $4.62$ & $5.13$ & $5.71$ & $6.41$ & $7.05$ & $6.64$ \\
                & & WA-PC & $5.45$ & $4.13$ & $4.49$ & $4.96$ & $5.29$ & $5.75$ & $6.38$ \\
                & & WA-PCC & $5.20$ & $3.76$ & $3.61$ & $3.74$ & $3.94$ & $4.22$ & $5.88$ \\
                & & DE & $4.12$ & $2.91$ & $3.36$ & $3.74$ & $4.19$ & $4.59$ & $4.65$ \\
                & & BMA & $4.72$ & $3.40$ & $3.86$ & $4.47$ & $5.04$ & $5.62$ & $5.76$ \\ \cline{3-10}
                & & \textbf{Avg} & $\mathbf{4.50}$ & $\mathbf{3.93}$ & $\mathbf{4.26}$ & $\mathbf{4.74}$ & $\mathbf{5.24}$ & $\mathbf{5.76}$ & $\mathbf{6.03}$ \\ \cline{2-10}
                & \multirow{6}{*}{\rotatebox[origin=c]{90}{\textbf{Recall}}} & UW-PC & $4.38$ & $4.82$ & $5.22$ & $5.82$ & $6.67$ & $7.31$ & $6.87$ \\
                & & UW-PCC & $4.14$ & $4.70$ & $5.22$ & $5.69$ & $6.54$ & $7.08$ & $6.63$ \\
                & & WA-PC & $3.78$ & $4.10$ & $4.52$ & $4.86$ & $5.33$ & $5.70$ & $6.39$ \\
                & & WA-PCC & $3.54$ & $3.75$ & $3.72$ & $3.80$ & $4.04$ & $4.24$ & $5.90$ \\
                & & DE & $2.37$ & $2.82$ & $3.38$ & $3.80$ & $4.27$ & $4.57$ & $4.60$ \\
                & & BMA & $2.95$ & $3.40$ & $3.95$ & $4.50$ & $5.10$ & $5.59$ & $5.78$ \\  \cline{3-10}
                & & \textbf{Avg} & $\mathbf{3.53}$ & $\mathbf{3.93}$ & $\mathbf{4.33}$ & $\mathbf{4.75}$ & $\mathbf{5.32}$ & $\mathbf{5.75}$ & $\mathbf{6.03}$ \\ \cline{2-10}
                & \multirow{6}{*}{\rotatebox[origin=c]{90}{\textbf{F1-Score}}} & UW-PC & $3.06$ & $4.67$ & $5.13$ & $5.71$ & $6.67$ & $7.23$ & $6.88$ \\
                & & UW-PCC & $4.97$ & $4.55$ & $5.13$ & $5.59$ & $6.54$ & $7.10$ & $6.64$ \\
                & & WA-PC & $4.61$ & $4.06$ & $4.40$ & $4.86$ & $5.31$ & $5.63$ & $6.26$ \\
                & & WA-PCC & $4.37$ & $3.70$ & $3.68$ & $3.78$ & $3.99$ & $4.19$ & $5.90$ \\
                & & DE & $3.30$ & $2.75$ & $3.32$ & $3.78$ & $4.23$ & $4.55$ & $4.69$ \\
                & & BMA & $3.89$ & $3.34$ & $3.92$ & $4.37$ & $5.07$ & $5.50$ & $5.77$ \\  \cline{3-10}
                & & \textbf{Avg} & $\mathbf{4.03}$ & $\mathbf{3.84}$ & $\mathbf{4.26}$ & $\mathbf{4.68}$ & $\mathbf{5.30}$ & $\mathbf{5.70}$ & $\mathbf{6.02}$ \\ \hline

            \end{tabular}
        \caption{Improvement attained in macro-averaged precision, recall, and F1-score by proposed MIP approach over other weighting schemes (D4).}
        \label{tab:D4Improvement}
        \end{table}
    \end{minipage}
\end{minipage}
\end{figure*}

Figures \ref{fig:D1Improvement}--\ref{fig:D4Improvement} demonstrate the percentage improvement achieved in balanced accuracy (primary metric) by the proposed MIP weighting scheme over all existing approaches for datasets D1--D4, respectively, for $K=2$ to $K=8$ classifiers selected from set $\mathcal{C}$. 
Tables \ref{tab:D1Improvement}--\ref{tab:D4Improvement} showcase the percentage improvement attained with respect to the secondary metrics (macro-averaged precision, recall, and F1-score) in each case. 
The average improvement per dataset is also shown in Table \ref{tab:avgImprovement}.  
Notably, MIP consistently outperformed the other techniques across all evaluation metrics, ensemble sizes, and datasets, demonstrating superior generalization and robustness.
Specifically, MIP yielded an increase in balanced accuracy ranging from 0.99\% to 7.31\%, with an overall average of 4.53\% across all datasets and ensemble sizes. Moreover, it attained an overall average increase of 4.63\%, 4.60\%, and 4.61\% in macro-averaged precision, recall, and F1-score, respectively.
In addition to these metrics, MIP also outperformed all other schemes with respect to the macro-averaged area under the precision-recall curve (AUPRC), as detailed in the provided supplementary material.


Furthermore, it can be observed that the improvement provided in all performance metrics by our approach over the other schemes generally increased as the ensemble size $K$ increased.
This reveals that MIP benefits more from a larger number of classifiers compared to the other methods. 
By optimally assigning weights to classifier-class pairs, our approach enhances the ability of the ensemble to leverage the diverse capabilities of each classifier, leading to a better performance as the number of classifiers grows.
On the other hand, the performance improvement was lower for the highly imbalanced datasets D1 and D2 (which had a higher imbalance ratio), compared to the moderately imbalanced datasets D3 and D4 (which had a lower imbalance ratio). This highlights the negative impact of class imbalance on the ensemble performance under all weighting schemes.
However, as demonstrated in Table \ref{tab:avgImprovement}, the performance gains for the highly imbalanced datasets were still significant and comparable to those observed for the moderately imbalanced datasets.
The effectiveness and robustness of our method under different cases of class imbalance are further demonstrated through the ablation studies included in the supplementary material. 
As reported by the Gurobi solver, the number of variables and constraints required for each dataset by our MIP approach are shown in Table \ref{tab:avgImprovement}.

\subsection{Weight Assignment Example}
\label{subsec:caseStudy}

To demonstrate the effective utilization of the strengths of each classifier in $\mathcal{C}$ by our method when assigning weights to classifier-class pairs, we consider the mean validation accuracy matrix $\mathbf{V}$ (given in Table \ref{tab:valAccuracyD2}) as obtained for dataset D2 and ensemble size $K = 8$. Although in the specific example classifier SVM (highlighted in gray) was not among the best-performing classifiers on average, it performed well for classes $A_2$ and $A_4$ (despite $A_4$ being among the minority classes in D2). Table \ref{tab:SVMWeights} shows the weights assigned to each class for SVM by each weighting approach. It can be observed that MIP (highlighted in gray) assigned relatively high weights for classes $A_2$ and $A_4$ where SVM performed well, in contrast to all other methods, which assigned significantly lower weights. On the other hand, MIP assigned low weights to the classes where SVM did not perform well.
Evidently, MIP leveraged the capabilities of SVM more effectively than all other approaches, even though SVM was among the lower-performing classifiers on average.

\subsection{Computational Efficiency}
\label{subsec:MIPExecutionEfficiency}

\begin{figure}[t]
    \centering
    \includegraphics[width=0.66\columnwidth]{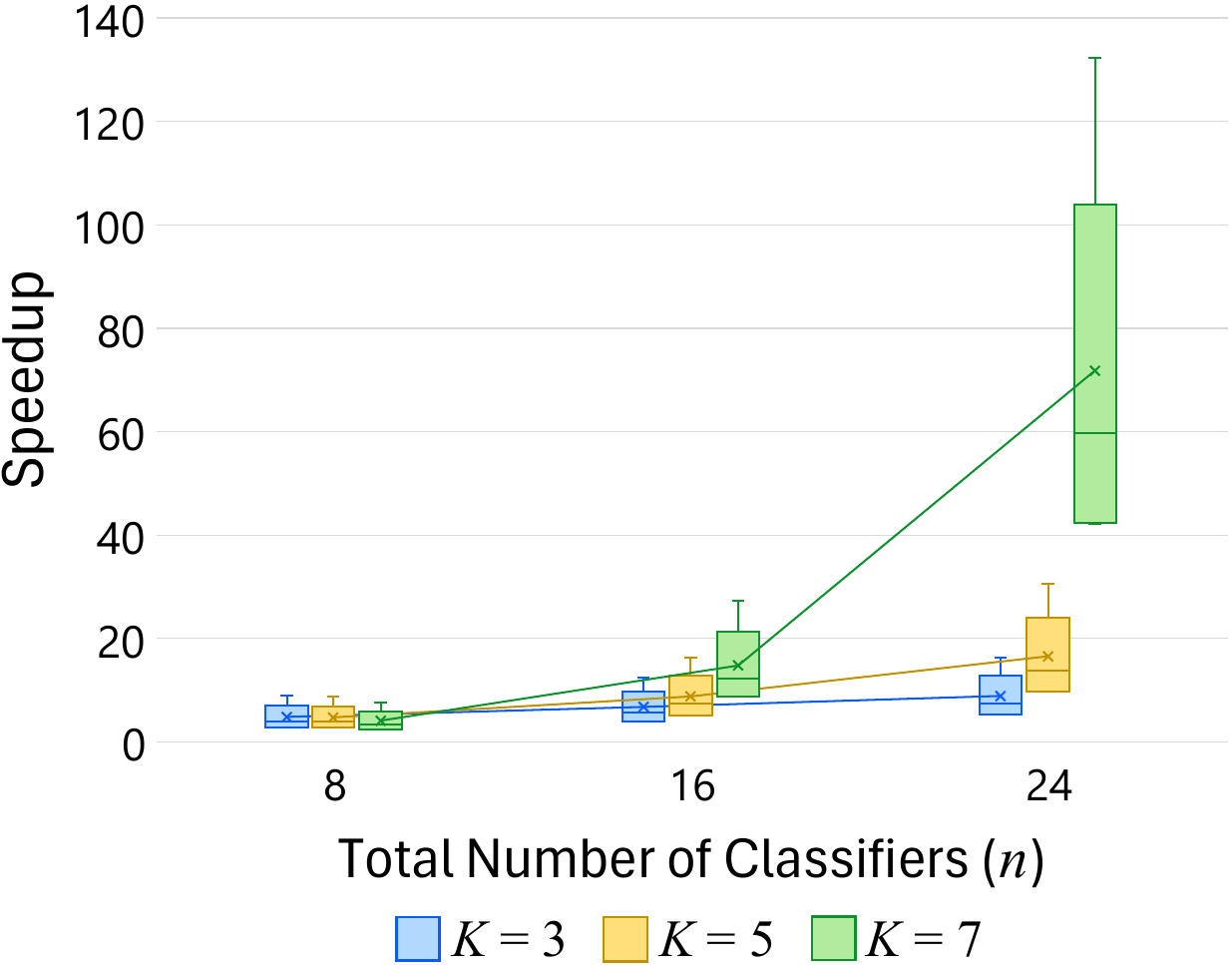}
    \caption{Speedup attained by proposed MIP approach over other schemes in calculating the weights of $K = \{3, 5, 7\}$ classifiers from a set of $n = \{8, 16, 24\}$, for dataset D2.}
    \label{fig:time_efficiencyMIPoverOthers}
\end{figure}

In addition to the significant performance gains provided by our method, MIP demonstrated computational efficiency in calculating the weights for $K$ classifiers from a set of $n$, compared to the other weighting schemes, despite the NP-hard nature of the examined problem. 
MIP seamlessly selects $K$ classifiers while optimally calculating their weights in a single run. In contrast, the other examined approaches calculate the weights for each possible classifier combination individually (i.e., for a total of $\binom{n}{K}$ combinations) before eventually selecting the $K$ best-performing classifiers.
Figure \ref{fig:time_efficiencyMIPoverOthers} illustrates the speedup achieved in weight calculation by MIP over the other methods, for dataset D2, $K=\{3, 5, 7\}$, and $n=\{ 8, 16, 24\}$.
It can be observed that the speedup provided by MIP increased significantly as $n$ grew, especially for larger ensemble sizes, highlighting its practicality compared to the other techniques. 
Even for smaller $K$ and $n$, MIP was considerably more efficient than the other weighting schemes.
For example, for $K=3$ and $n=8$, MIP required 0.35\,s to calculate the classifier weights, whereas the other approaches required 1 to 3.12\,s.

\begin{table}[t]
    \centering
    \fontsize{9}{10}\selectfont
    \begin{tabular}{lrccccc}
        \hline
        \multirow{2}{*}{$\mathcal{E}$} & \multicolumn{1}{c}{\multirow{2}{*}{$\rho$}} &  \textbf{\# Var/} & \multicolumn{4}{c}{\textbf{Avg. Increase (\%)}}\\ \cline{4-7}
        & &  \textbf{Constr.} & \textbf{Bal. Acc.} & \textbf{Precision} & \textbf{Recall} & \textbf{F1-score}\\
        \hline
        D1	& 6059.97 & 64/32 & 4.45	& 4.53	& 4.55	& 4.51 \\
        D2	& 1222.64 & 48/28 & 3.54	& 3.55	& 3.56	& 3.59 \\
        D3	& 5.91    & 40/26 & 5.42	& 5.52	& 5.50	& 5.51 \\
        D4	& 2.13    & 40/26 & 4.72	& 4.92	& 4.81	& 4.83 \\ \hline
        \multicolumn{2}{l}{\textbf{Overall}}  &     & \textbf{4.53} & \textbf{4.63} & \textbf{4.60} & \textbf{4.61} \\
        \hline
    \end{tabular}
    \caption{Average improvement achieved in all performance metrics by proposed MIP approach over other weighting schemes across all ensemble sizes for datasets D1--D4.} 
    \label{tab:avgImprovement}
\end{table}

\begin{table}[t]
    \centering
    \fontsize{9}{10}\selectfont
    \begin{tabular}{l|ccccc|c}
    \hline
        \multirow{2}{*}{$C_i$} & \multicolumn{5}{c|}{$E_j$} & \multirow{2}{*}{\textbf{Avg}} \\
        \cline{2-6}
         & $N_1$ & $A_1$ & $A_2$ & $A_3$ & $A_4$ &\\ \hline 
        MLR & 0.96 & 0.92 & 0.86 & 0.99 & 0.95 & 0.94 \\ 
        J48 & 0.89 & 0.78 & 0.85 & 0.90 & 0.90 & 0.86 \\ 
        JRIP & 0.90 & 0.74 & 0.78 & 0.89 & 0.96 & 0.85 \\ 
        REPTree & 0.76 & 0.86 & 0.80 & 0.98 & 0.73 & 0.83 \\ 
        MLP & 0.90 & 0.92 & 0.81 & 0.71 & 0.79 & 0.83 \\ 
        \rowcolor{gray!25} 
        SVM & 0.76 & 0.73 & \textbf{0.89} & 0.76 & \textbf{0.94} & 0.82 \\ 
        GNB & 0.90 & 0.85 & 0.81 & 0.71 & 0.73 & 0.80 \\ 
        IBk & 0.90 & 0.72 & 0.75 & 0.74 & 0.71 & 0.76 \\ \hline
    \end{tabular}
    \caption{Mean validation accuracy matrix $\mathbf{V}$ for dataset D2 and ensemble size $K=8$.}
  \label{tab:valAccuracyD2}
\end{table}

\begin{table}[!t]
    \centering
    \fontsize{9}{10}\selectfont
    \begin{tabular}{l|ccccc}
    \hline
        \multicolumn{1}{c|}{\textbf{Weighting}} & \multicolumn{5}{c}{$E_j$} \\ 
        \cline{2-6}
        \multicolumn{1}{c|}{\textbf{Scheme}} & $N_1$ & $A_1$ & $A_2$ & $A_3$ & $A_4$\\ \hline 
        UW-PC & 0.13 & 0.13 & 0.13 & 0.13 & 0.13 \\ 
        UW-PCC & 0.03 & 0.03 & 0.03 & 0.03 & 0.03 \\ 
        WA-PC & 0.11 & 0.11 & 0.11 & 0.11 & 0.11 \\ 
        WA-PCC & 0.02 & 0.02 & 0.03 & 0.02 & 0.03 \\ 
        DE & 0.12 & 0.12 & 0.12 & 0.12 & 0.12 \\ 
        BMA & 0.02 & 0.02 & 0.03 & 0.02 & 0.02 \\
        \rowcolor{gray!25}
        MIP & 0.00 & 0.03 & \textbf{0.20} & 0.04 & \textbf{0.23} \\ 
        \hline
    \end{tabular}
   \caption{Weights assigned to the SVM classifier (shown per class) by each weighting approach, for dataset D2 and ensemble size $K=8$.}
\label{tab:SVMWeights}
\end{table}

\section{Conclusion}
\label{sec:Conclusion}

We proposed an optimal, adaptable, and efficient MIP-based ensemble weighting scheme for imbalanced multi-class datasets to improve rare event detection in critical CPS.
Our method optimizes classifier weights on a granular per class basis, while seamlessly and optimally selecting a predefined number of classifiers from a given set.
Furthermore, it employs elastic net regularization to enhance robustness and generalization in weight assignment, while leveraging the strengths of each classifier.   
We compared our approach against six widely used weighting schemes, considering relevant datasets with multiple classes and varied imbalance ratios, under a wide range of ensemble sizes.
The experimental results demonstrate the superior performance and efficiency of our technique over all existing methods, attaining an increase in balanced accuracy ranging from 0.99\% to 7.31\%, with an overall average of 4.53\% across all datasets and ensemble sizes. 
Moreover, it achieved an overall average improvement of 4.63\%, 4.60\%, and 4.61\% in macro-averaged precision, recall, and F1-score, respectively, while also exhibiting computational efficiency.

\section{Acknowledgments}
This work has been supported by the European Union’s Horizon 2020 research and innovation programme under grant agreement No. 739551 (KIOS CoE) and from the Government of the Republic of Cyprus through the Cyprus Deputy Ministry of Research, Innovation and Digital Policy.

\bibliography{main}
\end{document}


\maketitle

\section{Ablation Studies}
\label{sec:ablation}
To gain deeper insights into the impact of class imbalance on the performance of the proposed MIP approach, we conducted two ablation studies using undersampling and oversampling techniques to generate varying class distributions for dataset D1. 
This dataset was selected due to its inherently high imbalance ratio $\rho$ and relatively large number of classes (compared to the other examined datasets), making it suitable for a broad range of modifications to its class distribution. 
In these studies, we evaluated the performance of our approach under: (a) an increasing imbalance ratio, and (b) an increasing number of minority classes in step imbalance scenarios.  
These cases represent diverse imbalance patterns commonly observed in multi-class datasets. Both ablation studies were conducted for ensemble size $K=8$.

\subsection{Ablation Study 1}
\label{subsec:ablationStudy1}
\begin{figure*}
    \centering
    \begin{minipage}[b]{0.58\textwidth}
        \begin{table}[H]
            \centering
            \fontsize{9}{10}\selectfont
            \begin{tabular}{r|ccccccc|c}
                \hline
                \multicolumn{1}{c|}{\multirow{2}{*}{$\rho$}} & \multicolumn{7}{c|}{\textbf{Class Distribution (\# Instances, \%)}} & \multicolumn{1}{c}{\multirow{2}{*}{\textbf{Total}}}\\
                \cline{2-8}
                                        & $N_1$ & $N_2$ & $N_3$ & $F_1$ & $F_2$ & $F_3$ & $F_4$ & \\
                \hline
        
                \multirow{2}{*}{\textbf{1.00}}   & 79\,700 & 79\,700 & 79\,700 & 79\,700 & 79\,700 & 79\,700 & 79\,700 & 557\,900\\
                                        & 14.286 & 14.286 & 14.286 & 14.286 & 14.286 & 14.286 & 14.286 & 100\%\\
                \hline                    
        
                \multirow{2}{*}{\textbf{3029.89}} & 17\,517 & 448\,423 & 88\,098 & 1\,721 & 1\,181 & 812 & 148 & 557\,900\\
                                         & 3.140 & 80.377 & 15.791 & 0.308 & 0.212 & 0.146 & 0.027 & 100\%\\
        
                \hline 
                
                \multirow{2}{*}{\textbf{6059.97}} & 17\,520 & 448\,438 & 88\,154 & 1\,721 & 1\,181 & 812 & 74 & 557\,900\\
                                         & 3.140 & 80.380 & 15.801 & 0.308 & 0.212 & 0.146 & 0.013 & 100\%\\
        
                \hline
        
                \multirow{2}{*}{\textbf{12121.95}} & 17\,520 & 448\,512 & 88\,117 & 1\,721 & 1\,181 & 812 & 37 & 557\,900\\
                                          & 3.140 & 80.393 & 15.794 & 0.308 & 0.212 & 0.146 & 0.007 & 100\%\\
            
                \hline
            \end{tabular}
            \caption{Imbalance ratio $\rho$ and corresponding class distribution for different versions of dataset D1 (ablation study 1).}
            \label{tab:ablationStudy1}
        \end{table}
    \end{minipage}
    \hfill
    \begin{minipage}[b]{0.38\textwidth}
        \centering
        \includegraphics[width=0.85\textwidth]{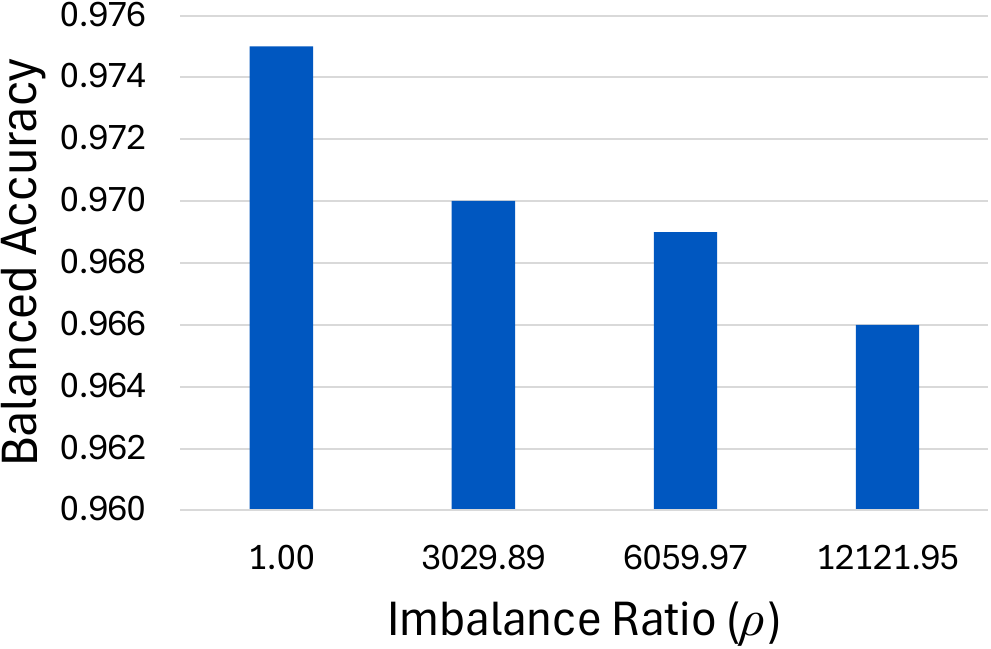}
        \caption{Balanced accuracy of proposed MIP approach under increasing $\rho$ (ablation study 1).}
        \label{fig:ablationStudy1}
    \end{minipage}
\end{figure*}

We generated resampled versions of dataset D1 with different imbalance ratios $\rho$. Specifically, we applied random undersampling to reduce the number of instances in classes exceeding a predefined target size and random oversampling with replacement to increase the number of instances in classes below their target size, while preserving the same total number of instances in the dataset.
The original dataset D1 has imbalance ratio $\rho = 6059.97$. The resampled version with $\rho = 1.00$ represents a baseline balanced dataset, while $\rho = 3029.89$ and $\rho = 12121.95$ correspond to resampled versions with approximately half and double the imbalance ratio of the original D1, respectively.
The class distribution for all versions of D1 is shown in Table \ref{tab:ablationStudy1}.
Figure \ref{fig:ablationStudy1} illustrates the balanced accuracy achieved by our MIP approach in each case.
It can be observed that the performance degraded slightly as $\rho$ increased. However, compared to the balanced dataset ($\rho=1.00$), the decrease in balanced accuracy was not significant, ranging from 0.51\% (for $\rho = 3029.89$) to 0.92\% (for $\rho = 12121.95$), highlighting the applicability and robustness of our approach to substantial variations in the imbalance ratio.

\subsection{Ablation Study 2}
\label{subsec:ablationStudy2}
\begin{figure*}
    \centering
    \begin{minipage}[b]{0.61\textwidth}
        \begin{table}[H]
            \centering
            \fontsize{9}{10}\selectfont
            \begin{tabular}{r|c|ccccccc|c}
                \hline
                \multicolumn{1}{c|}{\multirow{2}{*}{$r$}} & \multicolumn{1}{c|}{\multirow{2}{*}{$\rho$}} & \multicolumn{7}{c|}{\textbf{Class Distribution (\# Instances, \%)}} & \multicolumn{1}{c}{\multirow{2}{*}{\textbf{Total}}}\\
                \cline{3-9}
                                        &  & $N_1$ & $N_2$ & $N_3$ & $F_1$ & $F_2$ & $F_3$ & $F_4$ & \\
                \hline
        
                \multirow{2}{*}{\textbf{1}} & \multirow{2}{*}{6198.73} & 92\,981 & 92\,981 & 92\,981 & 92\,981 & 92\,981 & 92\,981 & 15 & 557\,901\\
                                                              & & 16.666 & 16.666 & 16.666 & 16.666 & 16.666 & 16.666 & 0.003 & 100\%\\
                \hline                    

                \multirow{2}{*}{\textbf{3}} & \multirow{2}{*}{6063.39} & 139\,458 & 139\,458 & 139\,458 & 139\,458 & 23 & 23 & 23 & 557\,901\\
                                                              & & 24.997 & 24.997 & 24.997 & 24.997 & 0.004 & 0.004 & 0.004 & 100\%\\
                \hline

                \multirow{2}{*}{\textbf{6}} & \multirow{2}{*}{6058.13} & 557\,348 & 92 & 92 & 92 & 92 & 92 & 92 & 557\,900\\
                                                              & & 99.901 & 0.016 & 0.016 & 0.016 & 0.016 & 0.016 & 0.016 & 100\%\\

                \hline
            \end{tabular}
            \caption{Number of minority classes $r$ and corresponding imbalance ratio $\rho$ and class distribution for different versions of dataset D1 (ablation study 2).}
            \label{tab:ablationStudy2}
        \end{table}
    \end{minipage}
    \hfill
    \begin{minipage}[b]{0.37\textwidth}
        \centering
        \includegraphics[width=0.85\textwidth]{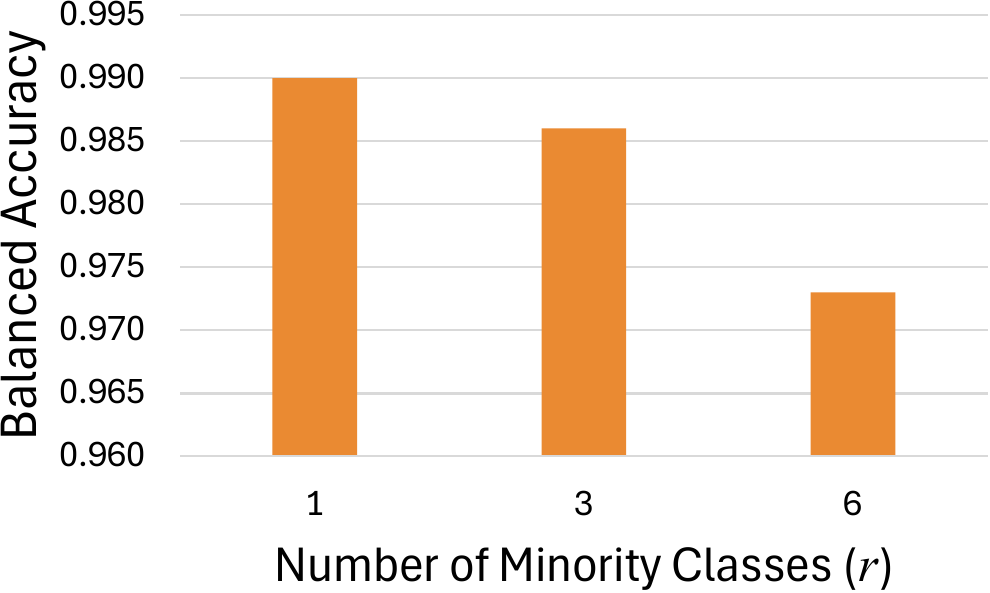}
        \caption{Balanced accuracy of proposed MIP approach under increasing $r$ (ablation study 2).}
        \label{fig:ablationStudy2}
    \end{minipage}
\end{figure*}

We simulated various step imbalance scenarios, setting the number of minority classes $r$ to one, three, and six out of the seven total classes in the dataset.
Specifically, we modified the class distribution of dataset D1 using resampling, similar to ablation study 1. In this study, all minority classes were assigned an equal number of instances $y$, and all majority classes were assigned an equal number of instances $z$, where $z \neq y$, creating a stepped imbalance between the two groups of classes while maintaining approximately the same imbalance ratio and dataset size.
The resulting class distribution in each scenario is shown in Table \ref{tab:ablationStudy2}.
Figure \ref{fig:ablationStudy2} depicts the balanced accuracy attained by the proposed MIP approach in each scenario.
It can be observed that the performance slightly decreased as the number of minority classes $r$ increased. The achieved balanced accuracy remained high, ranging from 0.990 (for $r=1$) to 0.973 (for $r=6$), demonstrating the effectiveness of our approach in rare event detection in imbalanced multi-class datasets.

\section{Additional Experimental Results}
To further evaluate the performance of the proposed MIP approach against the six other examined weighting schemes (UW-PC, UW-PCC, WA-PC, WA-PCC, DE, and BMA), we employed the macro-averaged area under the precision-recall curve (AUPRC) metric. Specifically, for each weighting scheme, we used the one-vs-rest approach to calculate the AUPRC for each class. In this approach, each class was treated as the positive class, while all other classes were combined into a single negative class by binarizing the true class labels. Thresholds were set based on the weighted predicted probabilities of the ensemble for each class. By varying these thresholds, we derived the precision-recall curve and calculated the corresponding AUPRC for each class using the trapezoidal rule. The macro-averaged AUPRC was subsequently obtained by averaging the AUPRC values across all classes.

\begin{figure*}[t]
\centering
   \begin{minipage}[b]{\columnwidth}
        \centering
        \includegraphics[width=0.97\columnwidth]{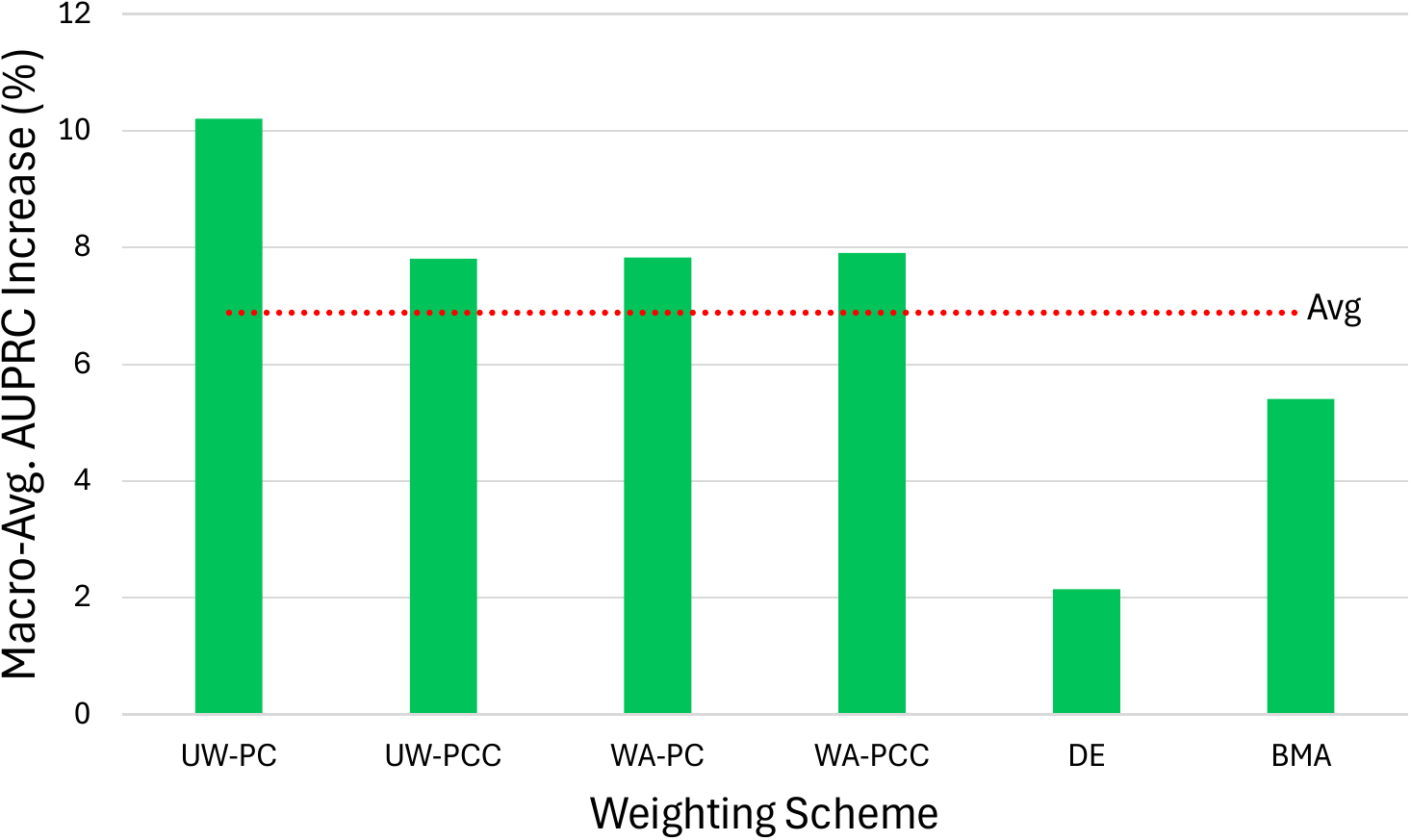}
        \caption{Improvement achieved in macro-averaged AUPRC by proposed MIP approach over other weighting schemes.} 
        \label{fig:AUPRC}
    \end{minipage}
\end{figure*}

Figure \ref{fig:AUPRC} illustrates the percentage improvement achieved in macro-averaged AUPRC by the proposed MIP approach over all other weighting schemes, for dataset D2 and ensemble size $K=8$. It can be observed that MIP outperformed all existing methods, providing an increase in macro-averaged AUPRC ranging from 2.14\% to 10.20\%, with an average of 6.88\% across all schemes.
Notably, the improvement in macro-averaged AUPRC exhibited a similar pattern to the improvement observed in the other evaluation metrics (balanced accuracy and macro-averaged precision, recall, and F1-score) for the specific dataset and ensemble size.

